# MATHEMATICAL MODEL FOR TRANSFORMATION OF SENTENCES FROM ACTIVE VOICE TO PASSIVE VOICE

Rakesh pandey, Himanshu Bahaguna and H.S.Dhami*
Department of Mathematics,
University of Kumaun,
SSJ Campus Almora,
Almora (Uttarakhand) INDIA-263601

## ABSTRACT

Formal work in linguistics has both produced and used important mathematical tools. Motivated by a survey of models for context and word meaning, syntactic categories, phrase structure rules and trees, an attempt is being made in the present paper to present a mathematical model for structuring of sentences from active voice to passive voice, which is is the form of a transitive verb whose grammatical subject serves as the patient, receiving the action of the verb.

For this purpose we have parsed all sentences of a corpus and have generated Boolean groups for each of them. It has been observed that when we take constituents of the sentences as subgroups, the sequences of phrases form permutation groups. Application of isomorphism property yields permutation mapping between the important subgroups. It has resulted in a model for transformation of sentences from active voice to passive voice. A computer program has been written to enable the software developers to evolve grammar software for sentence transformations.

## INTRODUCTION

The researchers at computer Science Department at Stanford University had worked on development of computer programs for accepting and manipulating transformational grammars corresponding to a version of the theory based on Chomsky's aspects of the theory of syntax. This project (1968) has made some interesting contributions to linguistic theory, particularly in the area of formal definitions of grammars, lexical insertion and traffic rules for transformations. In an attempt to obtain an effective rule for distinguishing sentences from non-sentences, Bargelli and Lambek (2001) have discussed the Mathematics of sentence structure. The arena of statistical analysis of texts has been applicable in information retrieval and natural language processing in general. In this context, we can cite the research paper of Dunning (1993), who has described the basis of a measure based on likelihood ratios for the analysis of a text. With the interest on categorical modeling of the cognitive abilities underlying the acquisition, use and understanding of natural language, Michael Moortgat (2002) has addressed two central questions of categorical grammar, the invariants of grammatical composition and the uniformity of the form/meaning correspondence across languages. Karin Muller (2002) has presented a probabilistic context free grammar, which describes the word and syllable structure of German words. Jakoboson (1961) had used mathematical approach in his book to define language and its structure in sentences.

Sentences can be written or spoken in the active or passive voice. In the active voice, the subject of the sentence acts upon something or someone. In the passive voice, the subject is acted upon. Active voice is distinguished from the passive voice by the identity of the actor. As a general rule, active voice is preferred because it meets two of the most important requirements of legal writing: clarity and conciseness. Active voice is clear because it focuses the reader's attention on the "doer of the action" it is also more concise simply because it usually involves fewer words. The passive voice is a very versatile construction. It is particularly useful when the performer of the action is unknown or irrelevant to the matter of the hand. Scientist ordinarily uses the passive voice to describe natural processes or phenomena under study. In technical and scientific articles, especially in the presentation of experimental methods, researchers use the passive voice as a convenient means of impersonal reporting. The passive voice allows them to avoid calling attention to themselves and to omit reference to any subjective thoughts or biases they might have bought to their work. The effect is to land the article the air of objectivity. In English as in many other languages, the passive voice is the form of a transitive verb whose grammatical subject serves as the patient, receiving the action of the verb. In transformation from an active-voice clause to an equivalent passive-voice construction, the direct object switch grammatical role. The direct object gets promoted to subject, and the subject demoted to an (optional) complement.

The book of Marcus Kracht (2003) contains an account of studies in language and linguistic theories from a mathematical point of view. Jean mark Gawron (2004) has defined the relations in linguistic aspects. The author has also generated some sets and relations for English Obstruents and has also suggested function approach for voice transformation in his book "Mathematical Linguistics".

This paper is an answer to the need of the students who use the English language as matter of course, but face difficulty in applying grammatical techniques in syntactic theory, like that of transformation of sentences from active voice to passive voice. Whether mathematics can have any answer to this language phenomenon, has been the concern of this paper.

## 1. MODUS OPERANDI

We define passivization of a sentence as-
Passivization: Transitive sentence $\longrightarrow$ Sentence.
Passivization: S $\longrightarrow$ the passive version of S.
As an illustration, we can cite the following example-
**Transitive:** John ate the bagel.
**Mapped to:** The bagel was eaten by John

We have made an attempt to transform a string '*a*' of active voice words into a string of passive voice words '*p*'. We are dealing it with the help of a mathematical tools based on algebraic properties.

## 2. STRUCTURING OF SENTENCES FROM ACTIVE VOICE TO PASSIVE VOICE

The basic sentence structure for the English language follows a ***SVO*** pattern, which means that the sentence begins with a subject (S) or something performing an action, followed by a verb (V) or the action, followed by an object (O) something that receives the action. The sentence in passive voice may be treated as the one having ***OVS*** pattern.

We can consider a set N defined as -

N = {i, we you, he, she, they, me, us, you, him, her, them, noun}  ..………(1.1)

Which can generate a cyclic group of subjects and objects order 13, as demonstrated in following table-

| I | We | You | He | She | They | Them | Her | Him | You | Us | Me | Noun |
|---|---|---|---|---|---|---|---|---|---|---|---|---|
| $a^1$ | $a^2$ | $a^3$ | $a^4$ | $a^5$ | $a^6$ | $a^{-6}$ | $a^{-5}$ | $a^{-4}$ | $a^{-3}$ | $a^{-2}$ | $a^{-1}$ | $a^0 = e$ |

**Table 1**

Since $a^1 * a^{-1} = e$ algebraically, so inverse of 'i' can be assumed to be 'me' and similarly for others.

Considering the set of articles

A = {the, a, an,………}  …………………..(1.2)

we can form an algebraic structure S for N and A in which N is closed with respect to an operation (say connection) with A, as

The – article, Policeman – subject ➔ the policeman – subject

Taking a set of all verbs according to their forms as

{V = { $v_{11}, v_{12}, v_{13}, v_{14}, v_{21}, v_{22}, v_{23}, v_{24}, v_{31}, v_{32}, v_{33}, v_{34}, v_{11}^{-1}, v_{12}^{-1}, v_{13}^{-1}, v_{14}^{-1}, v_{21}^{-1}, v_{22}^{-1}, v_{23}^{-1}, v_{24}^{-1}, v_{31}^{-1}, v_{32}^{-1}, v_{33}^{-1}, v_{34}^{-1}$ } ……………………….(1.3)

In $v_{ij}$ i represents tense and j represents form of tense.

Forms of actual elements for the verb 'write' can be depicted as under-

|  | **Verb form** | **Inverse element** |
|---|---|---|
| $v_{11}$ | Write | Written |
| $v_{12}$ | Is, am, are writing | Being written |
| $v_{13}$ | Have, has written | Been written |
| $v_{14}$ | Have been, has been written | ---------------- |
| $v_{21}$ | Wrote | Written |
| $v_{22}$ | Was, were writing | Being written |
| $v_{23}$ | Had written | Been written |
| $v_{24}$ | Had been written | ---------------- |
| $v_{31}$ | Shall, will Write | Be written |
| $v_{32}$ | Shall be, will be going | ---------------- |
| $v_{33}$ | Shall have, will have written | Have been written |
| $v_{34}$ | Shall have been, will have been written | ---------------- |

**Table 2**

We can now form an algebraic space with the help of three earlier defined structures S, V and O, where O is same as the group of subjects S.

Let  $a = (S)\#(V)*(O)$,  ……………..(1.4)

Where S stands for subject elements, V for verb elements and O for object elements. # and * represent respective exterior operations between them., for example in the sentence

*"The policeman has caught the thief."*

the structure $a$ can be permuted in 6 ways, which gives a new string of elements. Out of them one string is OVS.

Defining a transformation in-group S as

$$f(a^i) = (a^i)^{-1} \qquad \ldots\ldots\ldots\ldots\ldots\ldots(1.5)$$

we can take a mapping in set V

$$g(v_{ij}) = (v_{ij})^{-1} \qquad \ldots\ldots\ldots\ldots\ldots\ldots(1.6)$$

Since four elements of the set V, defined by (1.3), have not their inverses and thus it can be said that those elements map into null element. For these elements we form a new set named kernel of mapping, defined as

$$K(g) = \{v_{ij} : g(v_{ij}) = \phi\} = \{v_{14}, v_{24}, v_{32}, v_{34}\}. \qquad \ldots\ldots\ldots\ldots\ldots(1.7)$$

Transformation of algebraic structure $a$, defined by (1.4) shall be

$$T(a) = T(S_i \# V_{jk} * O_p) = (O_p^{-1} \otimes V_{jk}^{```} * S_i^{-1}) = p \qquad \ldots\ldots\ldots\ldots\ldots(1.8)$$

Where V``` represents the third form of verb.

Defining two exterior operations $\otimes$ and $*$, we can form a new algebraic structure

$$p = (O_p^{-1} \otimes V_{jk}^{```} * S_i^{-1}) \qquad \ldots\ldots\ldots\ldots(1.9)$$

which is not commutative under these operations.

In result (1.9), $\otimes$ is defined as

$$\otimes = A_{jk, p^{-1}} \qquad \ldots\ldots\ldots\ldots(1.10)$$

and $*$ is a conjunction operator defined between verb and subject elements.

The values of $A_{jk, p^{-1}}$ for particular objects and verbs defined in active voice shall be structured as elucidated in the following table-

| p →<br>ij ↓ | 1<br>I | 2<br>We | 3<br>You | 4<br>He | 5<br>She | 6<br>They | 7<br>Noun |
|---|---|---|---|---|---|---|---|
| 11 | Am | Are | Are | Is | Is | Are | Is, are |
| 12 | Am | Are | Are | Is | Is | Are | Is, are |
| 13 | Have | Have | Have | Has | Has | Have | Has, have |
| 21 | Was | Were | Were | Was | Was | Were | Was, were |
| 22 | Was | Were | Were | Was | Was | Were | Was, were |
| 23 | Had | Had | Had | Had | Had | Had | Had |
| 31 | Shall | Will | Will | Will | Will | Will | Will |
| 33 | Shall have | Will have | Will have | Will have | Will have | Will have | Will have |

**Table 3**

Equation (1.18) gives a rule by which active voice can transform into passive voice. This transformation is an exact rule for simple affirmative sentences. By taking an example we can easily conclude this. As,
  Active-: They are looking the movie.

In this sentence 'they' is subject by table 1 its value is $a^6$, 'are looking' is verb by table 2 its value is $v_{12}$ and the is article and movie is a noun but for we have earlier prove that the combination of article and noun is again in the category of subject/ object. Hence 'the movie' is an object. By table 1 inverse of $a^6$ is $a^7$, by table 2 is looking is mapped in being looked and again by table 1 inverse of 'the movie' is 'the movie'.

From equation (1.8), if we put the values in this we have,
$T(\boldsymbol{a}) = T(a_6 \# v_{12} * a_{13}) = (a_{13}^{-1} \otimes v_{12}^{```} * a_6^{-1}) = (a_{13} \otimes v_{12}^{```} * a_7) = \boldsymbol{p}$

Now from table 3, $\otimes$ can be finding.

Again retrieving sentence by giving the values to elements in transformation, we have
  Passive-: The movie is being looked by them.

## **Conclusion-:**

Our main aim for this paper to give an easier approach to understand sentence transformation. This work is an attempt to develop a software based on mathematical tolls, which is helpful to understand, how can a sentence change his form to another form without altering its sense. This work is in primary stage in which we have used only simple cases of voice transformation in affirmative sentences. We are looking in the direction in which we can give a wider range of transformation for all forms. We are also trying for all type of transformation as formation of compound, complex sentences and direct and indirect transformations.